\documentclass[10pt,twocolumn,letterpaper]{article}

\usepackage{iccv}
\usepackage{times}
\usepackage{epsfig}
\usepackage{graphicx}
\usepackage{amsmath}
\usepackage{amssymb}

\usepackage{algorithm}
\usepackage{algorithmic}

\usepackage{booktabs}
\usepackage{subcaption}
\usepackage{multirow}
\usepackage{dsfont}
\usepackage{enumitem}
\usepackage{soul}
\usepackage{bbm}

\usepackage[accsupp]{axessibility}

\usepackage[breaklinks=true,bookmarks=false]{hyperref}

\iccvfinalcopy 

\def\iccvPaperID{1798} 

\ificcvfinal\fi

\begin{document}

\title{MetaGCD: Learning to Continually Learn in Generalized Category Discovery}

\author{Yanan Wu$^{1,2}$\thanks{The authors contributed equally to this work.}, Zhixiang Chi$^{3}$\footnotemark[1], Yang Wang$^{4}$, Songhe Feng$^{1,2}$ \thanks{Corresponding Author}\\
$^{1}$Key Laboratory of Big Data \& Artificial Intelligence in Transportation, \\ Ministry of Education, Beijing Jiaotong University, Beijing, 100044, China\\
$^{2}$School of Computer and Information Technology, Beijing Jiaotong University, Beijing, 100044, China\\
$^{3}$Department of Electrical and Computer Engineering, University of Toronto, Toronto, M5G1V7, Canada\\
$^{4}$Department of Computer Science and Software Engineering,\\ Concordia University, Montreal, H3G2J1, Canada\\
{\tt\small \{ynwu0510,shfeng\}@bjtu.edu.cn}, {\tt\small zhixiang.chi@mail.utoronto.ca}, {\tt\small yang.wang@concordia.ca}
}

\maketitle
\ificcvfinal\thispagestyle{empty}\fi

\begin{abstract}
In this paper, we consider a real-world scenario where a model that is trained on pre-defined classes continually encounters unlabeled data that contains both known and novel classes. The goal is to continually discover novel classes while maintaining the performance in known classes. We name the setting Continual Generalized Category Discovery (C-GCD). 
Existing methods for novel class discovery cannot directly handle the C-GCD setting due to some unrealistic assumptions,
such as the unlabeled data only containing novel classes. Furthermore, they fail to discover novel classes in a continual fashion. In this work, we lift all these assumptions and propose an approach, called MetaGCD, to learn how to incrementally discover with less forgetting. Our proposed method uses a meta-learning framework and leverages the offline labeled data to simulate the testing incremental learning process. A meta-objective is defined to revolve around two conflicting learning objectives to achieve novel class discovery without forgetting. Furthermore, a soft neighborhood-based contrastive network is proposed to discriminate uncorrelated images while attracting correlated images. We build strong baselines and conduct extensive experiments on three widely used benchmarks to demonstrate the superiority of our method. Our code is available at \href{https://github.com/ynanwu/MetaGCD}{https://github.com/ynanwu/MetaGCD}.
\end{abstract}

\begin{figure}[t]
\centering
\centerline{\includegraphics[scale=0.47]{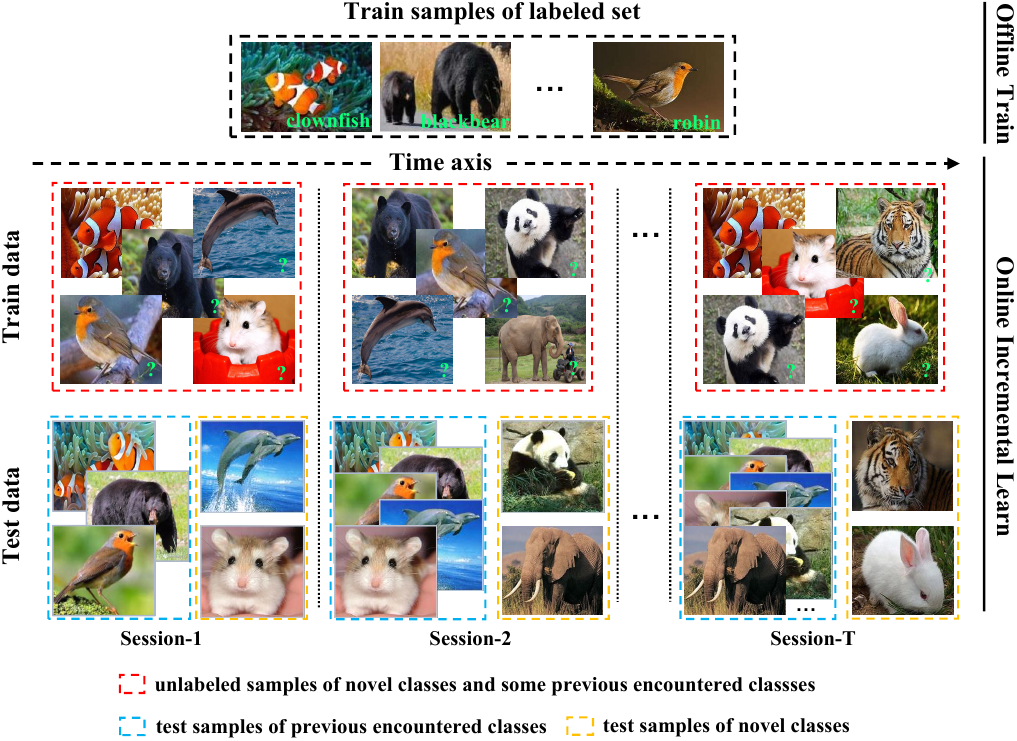}}
\caption{{\bf Illustration of our C-GCD setting}. During the offline training, we learn an initial model based on training samples of the labeled set. During each subsequent online incremental learning, we are given some unlabeled images belonging to both known and novel classes. Our goal is to update the model in each incremental session so that the model can maintain the performance on old classes while discovering novel classes.} 
\label{fig1}
\end{figure}

\section{Introduction}
Object categories in real-world environments are dynamically evolving and expanding over time. However, conventional deep learning-based visual recognition methods normally focus on closed-world scenarios with pre-defined categories~\cite{he2016deep,simonyan2014very}. Such systems are brittle when deployed to an ever-changing realistic open-world setting, where object instances may come from new categories. In contrast, recognizing the known categories and utilizing them to discern the unknowns are intrinsic to human perception. 

Recently, discovering the novel classes among unlabeled data has been an active area of research~\cite{han2020automatically,fini2021unified,vaze2022generalized,roy2022class,zhang2022grow}. However, most prior works make several assumptions that are unrealistic in practice. For example, the works in \cite{han2020automatically,fini2021unified,vaze2022generalized} assume the co-existence of both labeled data (with known classes) and unlabeled data (contains potential unknown classes to be discovered) at the training phase and the models are learned from scratch. This leads to repetitive large-scale training every time when new classes are expected to be discovered. The works in \cite{han2020automatically,fini2021unified,roy2022class,zhang2022grow} assume the newly encountered unlabeled data only belongs to the novel classes. This is unrealistic in practice. To meet such conditions, a rigorous filtering method is needed to precisely filter out known class data to avoid degenerate solutions. Due to these limitations, none of these works can be used to build recognition systems that can deal with evolving object categories sequentially over a long time horizon.  

In this paper, we consider a more flexible setting for real-world applications. Let us consider the application of home robots. The robots are equipped with an offline trained object recognition model on pre-defined categories during manufacturing. After deployment, the robots are expected to operate in diverse environments. While operating, they continually receive data that belongs to known and possibly unknown classes. Ideally, we would like the robots to continually discover and learn novel classes from such data. We dub such a setting as \textit{\textbf{Continuous Generalized Class Discovery (C-GCD)}}. As shown in Fig.~\ref{fig1}, C-GCD has two phases: 1) an offline training phase that allows the model to be trained on large-scale labeled data with pre-defined classes; 2) when the model is deployed, it continually encounters unlabeled data that comes from both known and novel classes on a \textit{longer horizon}. At each incremental session, the data from the previous sessions is \textit{inaccessible}. The model needs to precisely classify the known classes and discover novel ones to expand its knowledge base. Obviously, the main challenge of C-GCD is to discover the novel classes among \textit{unlabeled} images that contain both \textit{known} and \textit{unknown} categories while maintaining the performance on \textit{old} classes. However, learning novel knowledge normally leads to notorious catastrophic forgetting~\cite{chi2022metafscil}, which further exacerbates the model performance.

There are some initial attempts on C-GCD~\cite{joseph2022novel, zhang2022grow}. However, they only consider C-GCD at the deployment stage mentioned above. The offline training stage is not fully exploited in these works. Concretely, the labeled data during offline training is only used for pre-training model \textit{representations}. Therefore, the model at the offline stage is unaware of its subsequent learning duty (discover novel classes and retain the performance of known classes)~\cite{chi2021test} and is also prone to overfit to the labeled set~\cite{vaze2022generalized}. Such learning objective misalignment leads to cumbersome heuristic strategies to facilitate the new learning task while keeping the previous knowledge. For example, to learn the novel classes, ~\cite{joseph2022novel} requires a self-labeling method, which may cause error propagation. A routing strategy is also required to determine the known and novel classifier heads. 
\cite{zhang2022grow} relies on a thresholding method to filter novel instances.
However, the overall robustness of the method can be sensitive to the threshold. To alleviate the forgetting issues, \cite{joseph2022novel, zhang2022grow} propose to distill the knowledge from the pre-trained base models. Data replay is also utilized to either directly select representative labeled examplars~\cite{zhang2022grow} or generate pseudo-latent representations from them~\cite{joseph2022novel}. Consequently, the base models and the replay buffers have to be stored locally which may cause storage problems, especially in a resource-constraint environment. 

In this work, we propose a fully learning-based solution, named MetaGCD, to minimize the hand-engineered heuristics in prior works. Concretely, at the offline training phase, instead of pre-training a model \textit{representation}, we directly train an \textit{initialization} that is learned to discover novel categories with less forgetting when deployed. It is realized by meta-learning-based bi-level optimization~\cite{finn2017model} to couple the offline training and downstream learning objectives. During the offline training, we simulate the testing scenario and construct pseudo incremental novel class discovery sessions using the labeled data. At each incremental session, we discover novel classes by updating the model using an unsupervised contrastive loss. The meta-objective is then defined by validating the updated model on \textit{all} classes encountered on a labeled pseudo test set. Therefore, the meta-objective of the offline training is aligned with the evaluation protocol at deployment. It enforces the model to learn to balance two conflicting objectives, namely discovering new objects and not-forgetting old objects. The meta-objective also reinforces the unsupervised updated model to be supervised by the true labels to ensure valid novel class discovery.

MetaGCD uses unsupervised contrastive learning to explore the relationship among instances for novel class discovery. Therefore, it is less prone to label overfitting~\cite{vaze2022generalized}. However, we observe that the negative pairs in contrastive learning normally dominate the loss function. So we further propose soft neighborhood contrastive learning to mine more positiveness. Concretely, for each image instance, we select the nearest candidate neighbors within the batch to treat them as soft positive samples to contribute to the discriminative feature learning. Overall, our contributions are summarised as follows:
\begin{itemize}[noitemsep]
    \item We consider a realistic setting C-GCD for applications in real-world scenarios. It allows the model trained on pre-defined classes to continually explore novel classes through incoming unlabeled data while simultaneously keeping the performance of known classes.     
    \item We propose a meta-learning approach where the learning objective is well aligned with the evaluation protocol during testing. It directly optimizes the model to achieve novel object discovery without forgetting. 
    \item A soft neighborhood contrastive learning method is also proposed to mine more soft positive pairs to elevate the discovery capability.   
    \item We establish strong baselines and show that our method achieves superior performance with less hand-engineered design through extensive experiments.
\end{itemize}

\section{Related Work}

\noindent{\textbf{Discovering novel classes.}} \textit{Novel Class Discovery} (NCD) aims to discover the novel classes from unlabeled data by utilizing the prior knowledge from the labeled data~\cite{han2020automatically,fini2021unified,han2019learning}. However, NCD assumes that the unlabeled data only belongs to the novel classes, which is unrealistic. Alternatively, a \textit{generalized} version of NCD (GCD)~\cite{vaze2022generalized} relaxes such constrain. Although GCD allows the unlabeled data to contain both known and novel classes, they are both required to be present in the training phase. It leads to repetitive large-scale training when different groups of unlabeled data are continually presented to the recognition system. Recently, a class incremental variant of NCD (class-iNCD) is proposed to learn the tasks of labeled known and unlabeled novel classes sequentially~\cite{roy2022class}. When learning the novel classes, the data of old classes are inaccessible.  In the end, the model is evaluated on all encountered classes. 
Nevertheless, only a few incremental sessions containing unlabeled novel classes are allowed in class-iNCD. This limitation hinders its applicability under the realistic setting with continually evolved object categories. Our proposed C-GCD alleviates the above limitations in real-world scenarios. Our approach can learn from labeled pre-defined classes during offline training, and then continuously encounter unlabeled data with both known and novel classes after deployment. Our model will learn to discover novel classes without forgetting old classes. C-GCD is also related to the classic class-incremental setting~\cite{{li2017learning,wu2019large,liu2020mnemonics,liu2022few}}. But C-GCD is more challenging as the newly evolved classes are unlabeled and an automatic class discovery mechanism is required~\cite{joseph2022novel}. 

\noindent{\textbf{Meta-learning.}} Existing meta-learning methods can be categorized into: 1) model-based~\cite{santoro2016meta,bateni2020improved,chen2022bidirectional}; 2) Optimization-based~\cite{ravi2017optimization,finn2017model,chen2021generalized}; and 3) metric-based ~\cite{snell2017prototypical}. Typical meta-learning methods utilize bi-level optimization to train a model that is applicable for downstream adaptations. Our work is built upon MAML~\cite{finn2017model}, which trains a model initialization through episodes of tasks for fast adaptation via gradient updates. Such learning paradigm has been widely applied in different vision tasks, such as test-time adaptation~\cite{soh2020meta,liu2023meta,lu2020few,liu2022towards}, continual learning~\cite{zhang2021few,javed2019meta,wu2023metazscil} and domain shift~\cite{zhang2021adaptive,zhong2022meta,gu2022improving}. In our case, the adaptation is achieved in an unsupervised manner, and the bi-level optimization is utilized to combine two conflicting learning objectives: discovering the novel classes without forgetting the old classes.

\noindent{\textbf{Contrastive learning.}} Contrastive learning has been popular in self-supervised visual representation learning~\cite{bachman2019learning,chen2020simple,he2020momentum,pan2021videomoco,liang2022self}. It explores the relationships among data instances by constructing positive and negative pairs. Therefore, the overfitting on the label space is reduced to improve the generalization of downstream tasks. Zhong \textit{et al.}~\cite{zhong2021neighborhood} apply contrastive learning to discover novel classes by exploring the data neighborhood and choosing pseudo-positive pairs. However, those pseudo-positive pairs contribute equally regardless of their closeness compared to the reference sample. In this work, we introduce the soft positiveness concept to allow adaptive contribution. 

\begin{figure*}[t]
\centering
\includegraphics[width =0.80\linewidth]{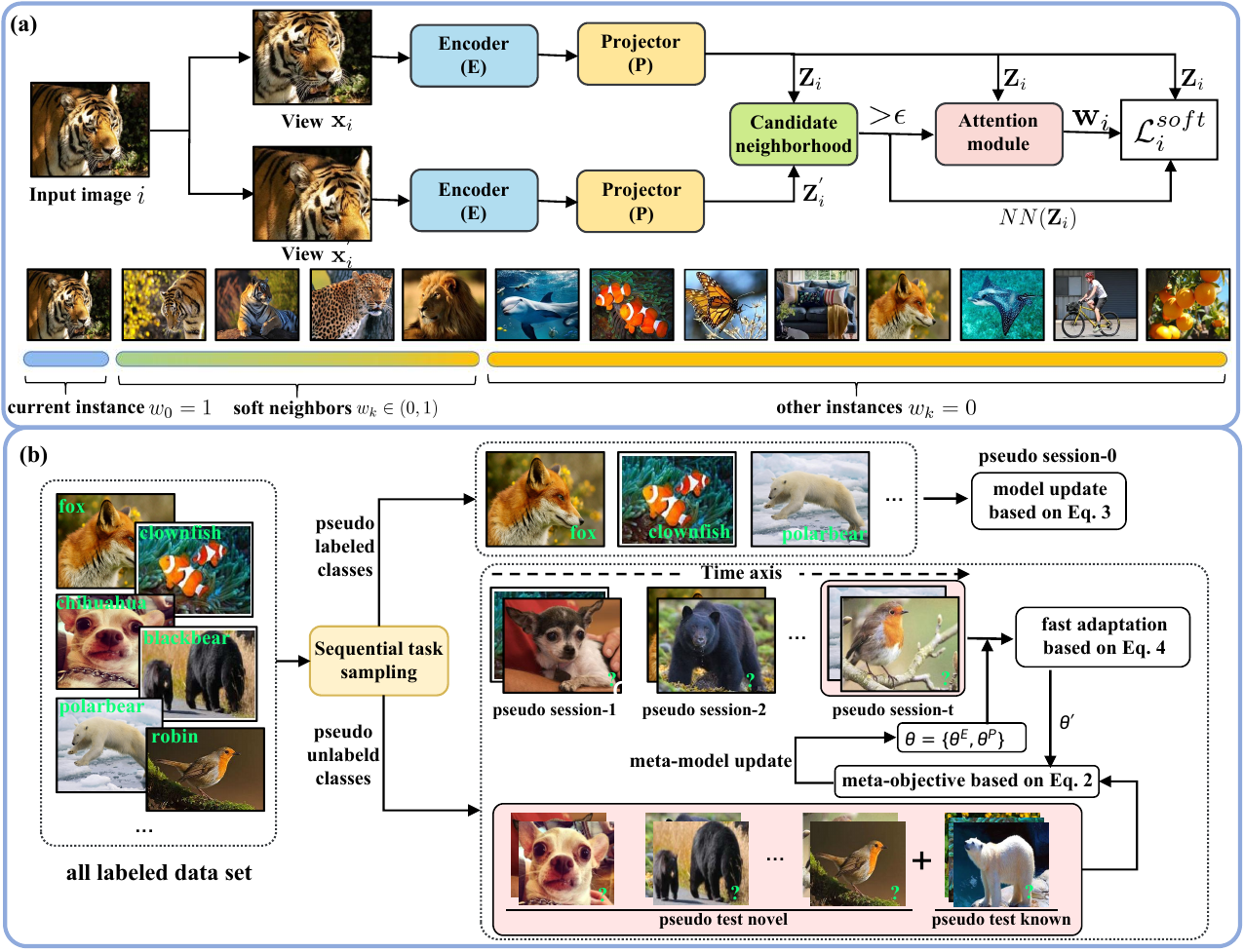} 
\caption{{\bf Overview of the proposed MetaGCD}. (a) Our soft neighborhood contrastive learning network aims to discriminate uncorrelated instances while absorbing correlated instances to learn discriminative representations. (b) Our meta-learning optimization strategy utilizes the offline labeled data to simulate the testing incremental learning process by sampling sequential learning tasks. By learning from these sampled sequential tasks, our model learns a good initialization, so that it can effectively adapt to discover new novel classes without forgetting old classes.} 
\label{fig2}
\end{figure*}

\section{The Proposed Method}

\paragraph{Problem definition.} The goal of C-GCD is to have the offline trained model continually discover and learn novel object classes from unlabeled data containing both known and novel classes. We define a sequence of $T$ learning sessions $\{\mathcal{S}^0, \mathcal{S}^1, \cdots, \mathcal{S}^T \}$. Let $x^t \in \mathcal{X}^t$ and $y^t \in \mathcal{Y}^t$ denote the input and label space at session $t$. We represent each session as: $\mathcal{S}^0 = \{(\mathbf{x}_i^0, \mathbf{y}_i^0)\}_{i=1}^{Z_0}$ and $\mathcal{S}^t = \{(\mathbf{x}_i^t)\}_{i=1}^{Z_t}$. Note, only the first session (\textit{i.e.}, $t=0$) contains large-scale labeled samples. As for $t>0$,  $\mathcal{S}^t$ only contains \textit{unlabeled} data. At the $t^{th}$ session, only $\mathcal{S}^t$ is accessible, and the incoming data belongs to both learned \textit{known} class from previous sessions and \textit{novel} classes. Therefore, we can denote $\mathcal{Y}^t = \mathcal{Y}^{t-1} \cup \mathcal{Y}^t_n$, where $\mathcal{Y}^t_n$ represents the \textit{novel} classes to be discovered at session $t$. After learning on $\mathcal{S}^t$, the model is evaluated on all test images accumulated until session $t$ to test the performance on $\mathcal{Y}^{t-1}$ (ideally the model should not forget old classes) and the discovery capability on $\mathcal{Y}^t_n$.
Compared with previous works~\cite{han2020automatically,fini2021unified,roy2022class,zhang2022grow}, C-GCD is much more challenging due to several factors. First, the unlabeled data contains both \textit{known} and \textit{unknown} classes,  \textit{i.e.}, $\mathcal{Y}^t = \mathcal{Y}^{t-1} \cup \mathcal{Y}^t_n$. Second, labeled data is absent at $t>0$, \textit{i.e.}, $\mathcal{S}^0 \cup \mathcal{S}^t = \varnothing$ where $t>0$. Finally, since C-GCD operates on a long horizon, \textit{i.e.}, $t \gg 1$, the catastrophic forgetting issue is more severe.

\noindent{\bf Method overview.} Fig.~\ref{fig2} shows an overview of MetaGCD. Following~\cite{vaze2022generalized}, we use a model without parametric classification heads since it is more suitable for dealing with novel classes. Novel class discovery is performed by directly clustering the feature spaces and class labels are assigned through the classic $k$-means algorithm. Concretely, we learn a model initialization using the labeled data during offline training. 
During each continual learning session, we update the model using a soft neighborhood contrastive learning~(see Fig.~\ref{fig2} (a)) on unlabeled data. To fully exploit the labeled data in offline learning, we further develop a bi-level optimization based on meta-learning to simulate the online learning scenario, so that the model is ready to adapt to new incoming unlabeled data and discover novel objects after the offline training (see Fig.~\ref{fig2} (b)). In the following, we describe these two parts of our method in detail. 

\subsection{Contrastive learning based clustering network} 
Considering the characteristics of labeled and unlabeled data, we employ different contrastive learning strategies. To train on the labeled data, we utilize a combination of unsupervised and supervised contrastive losses. When discovering the latent classes in continually encountered unlabeled data, we propose to mine soft positive neighbors for each data instance to elevate the discriminative feature learning. 

\subsubsection{Representation learning on labeled data} 
To learn a robust and semantically meaningful representation on labeled data, we utilize both self-supervised \cite{gutmann2010noise} and supervised~\cite{khosla2020supervised} contrastive losses. Let $\textbf{x}_i$ and $\textbf{x}^{\prime}_{i}$ be two randomly augmented versions of $i^{th}$ instance sample, the unsupervised contrastive loss is expressed as:
\begin{equation}
\mathcal{L}_i^{ucl}=-\log\frac{\exp(\textbf{z}_i\cdot{\textbf{z}_i^{\prime}/\tau})}{\sum_n \mathbb{I}_{[n\neq i]} \exp(\textbf{z}_i\cdot \textbf{z}_n/\tau)}
\label{uns_loss}
\end{equation}
where $\textbf{z}_i = \phi(f(\textbf{x}_i))$, $\mathbb{I}_{[n\neq i]}$ is an indicator function,
and $\tau$ is a temperature value. $f$ is the feature extractor, and $\phi$ is a multi-layer perceptron (MLP) projection head.

The supervised contrastive counterpart is expressed as:
\begin{equation}
\mathcal{L}_i^{scl}=-\frac{1}{|\mathcal{N}(i)|}\sum_{q\in \mathcal{N}(i)}\log\frac{\exp(\textbf{z}_i\cdot \textbf{z}_q/\tau)}{\sum_n \mathbb{I}_{[n\neq i]} \exp(\textbf{z}_i\cdot \textbf{z}_n/\tau)}
\label{s_loss}
\end{equation}
where $\mathcal{N}(i)$ denotes the indices of instances having the same label as $\textbf{x}_i$ within the batch. Finally, these two losses are weighted by $\lambda$ to train on the labeled data: 
\begin{equation}
\mathcal{L}_{labeled}=(1-\lambda)\sum_{i\in B}\mathcal{L}_i^{ucl}+\lambda\sum_{i\in B}\mathcal{L}_i^{scl}
\label{jointly}
\end{equation} 

\subsubsection{Soft neighborhood contrastive learning on unlabeled data}
When learning on unlabeled data, only the unsupervised contrastive loss Eq.~\ref{uns_loss} can be used. However, the samples within the same class could be mistakenly treated as negatives due to the missing labels. In addition, the number of negative pairs significantly surpasses positive pairs. Such imbalanced loss contribution could be sub-optimal. Aligning the positive and negative pairs with the true classes emerges as a desired solution. \cite{zhong2021neighborhood} has attempted to address the limitations by mining more positive pairs in the neighbored of each data sample. However, the pseudo-positive pairs are treated equally, regardless of how close they are to that data sample. To address this issue, we propose to encode soft positive correlation among instance neighbors to achieve adaptive contribution, as shown in Fig.~\ref{fig2}~(a). 

Specifically, for each $\textbf{x}_i$, we first use the nearest neighbor operator on the projected features to select candidate neighbors. We denote them as $NN(\textbf{z}_i)_k$ with $k$ as the index. We then pass them and $\textbf{z}_i$ to an attention module to predict a set of positiveness values $\mathbf{w}_{i} =  \{w_{ik}\} \in (0, 1)$ to weight the contribution of $NN(\textbf{z}_i)_k$ to the loss. Accordingly, the soft neighborhood contrastive loss is defined as:
\begin{equation}
\mathcal{L}_i^{soft}=-\frac{1}{|NN(\textbf{z}_i)|}\sum_{k\in NN(\textbf{z}_i)}\log\frac{w_{ik}\cdot \exp(\textbf{z}_i\cdot \textbf{z}_k/\tau)}{\sum_n \mathbb{I}_{[n\neq i]} \exp(\textbf{z}_i\cdot \textbf{z}_n/\tau)}
\label{eq:soft}
\end{equation}

\noindent{\textbf{Candidate neighborhood.}}
For each batch of data, we first compute their features $\mathcal{F}$ from the projection head at once. For each reference view $\textbf{x}_i$, we retrieve nearest neighbors by comparing the cosine similarity to a threshold $\epsilon$ as:
\begin{equation}
NN(\textbf{z}_i)=\{\mathbf{F}\}, \ \textrm{for} \ \mathbf{F} \ \textrm{in} \ \mathcal{F}, \ \textrm{if} \ \cos(\textbf{z}_i,\mathbf{F}) \geq \epsilon
\label{eq:nearest}
\end{equation}
where $\textbf{z}_i$ and $\mathbf{F} \in \mathcal{F}$ are normalized before computation. 

\noindent\textbf{Positiveness generation.} 
Fig.~\ref{fig2} (a) shows an intuitive example of the candidate neighbors. The first two neighbors belong to the same category as the reference `\textit{tiger}' sample, while the $3^{rd}$ and $4^{th}$ neighbors are partially related (\textit{i.e.}, they belong to the `\textit{lion}' and `\textit{leopard}' categories, but not `\textit{tiger}'). The remaining instances are not related to `\textit{tiger}'. Therefore, the first four instances tend to be selected and they should contribute adaptively to the loss. We propose to learn an attention module to measure the soft correlations between the selected neighbors and the reference instance (instead of the binary form in~\cite{zhong2021neighborhood}). Given two inputs $\textbf{z}_i$ and $NN(\textbf{z}_i)_k$, we can calculate the positiveness score as:
\begin{equation}
\mathbf{w}_{i}=\textrm{Softmax}[f_1(\textbf{z}_i) \times f_2(NN(\textbf{z}_i)_k)^T]
\label{weight}
\end{equation}
where $f_{1}(\cdot)$ and $f_{2}(\cdot)$ are the new projection layers and $\times$ denotes the cross attention operator. Eq.~\ref{weight} is then normalized so that $\mathbf{w}_i$ has a max value as $1$. Note that $f_{1}(\cdot)$ and $f_{2}(\cdot)$ can also be the non-parametric identity mappings, which are empirically found to be more effective. This observation may be attributed to the self-supervised learning paradigm, where the objective is to train the encoder effectively. Simplifying the attention module leads to less overfitting and improves learning attentive features.

\subsection{Learning to incrementally discover categories}
The main limitation of the prior works is that the labeled set $\mathcal{S}^0$ is not fully exploited~\cite{chi2022metafscil}. Instead of performing only representation learning on $\mathcal{S}^0$~\cite{joseph2022novel, zhang2022grow}, we borrow the meta-learning paradigm (in particular MAML~\cite{finn2017model}) to learn how to continually discover new classes. In few-shot learning, during meta-training, MAML constructs few-shot tasks to mimic the meta-testing scenario to achieve learning to quickly adapt. In our C-GCD case, the online continual class discovery tasks can be viewed as the ``meta-testing" stage. Therefore, we propose to simulate the continual setting using $\mathcal{S}^0$ during offline training, as shown in Fig.~\ref{fig2}~(b). We aim to produce a model \textit{initialization} that is trained by aligning the training and evaluation objectives so that it is endowed with the capability to effectively discover novel classes with less forgetting during evaluation.

\noindent\textbf{Sequential task sampling.} To mimic the evaluation process, we sample sequential learning tasks from $\mathcal{S}^0$~\cite{chi2022metafscil}. Specifically, we first randomly separate the $\mathcal{S}^0$ into \textit{pseudo labeled} and \textit{pseudo unlabeled classes} without overlapping. Next, we sample a sequence of $T+1$ sessions, $\mathcal{D}=\{(\mathcal{D}^j_{tr}, \mathcal{D}^j_{te})\}^T_{j=0}$, where $\mathcal{D}^j_{tr}$ and $\mathcal{D}^j_{te}$ are the training and test set for the $j^{th}$ session. 
For the training splits $\{\mathcal{D}^j_{tr}\}$, we follow the evaluation protocol to only allow the first session to contain labeled data, (\textit{i.e.}, $\mathcal{D}^0_{tr} = \{\mathbf{x}_{tr}^0, \mathbf{y}_{tr}^0\}$) and the rest with unlabeled data (\textit{i.e.}, $\mathcal{D}^{j}_{tr} = \{\mathbf{x}_{tr}^{j}\}$, for $j>0$).
We also set the first session to contain a larger number of samples, \textit{i.e.}, $\left | \mathcal{D}^0_{tr} \right | \gg \left | \mathcal{D}^{j>0}_{tr} \right |$ to simulate the C-GCD setting where the model is first trained during an offline training stage with a large amount of data. For the test splits $\{\mathcal{D}^j_{te}\}$, all of them contain labels that will be used during the optimization (\textit{ i.e.}, $\mathcal{D}^j_{te} = \{\mathbf{x}_{te}^j, \mathbf{y}_{te}^j\}$, $\forall j$). Note that $\mathcal{D}^j_{te}$ only contains the test data belonging to the current session $j$.

\begin{algorithm}[t]
\caption{The optimization procedure of MetaGCD}
\begin{algorithmic}
\label{Algorithm one}
\REQUIRE $\alpha$, $\beta$, $\gamma$: learning rates \\
\REQUIRE $\mathcal{S}^0$: training set of \textit{labeled} classes\\
\STATE 1: randomly initialize parameters $\theta$\\
\STATE 2: \textbf{while} not converged \textbf{do}\\
\STATE 3: \quad $\mathcal{D}$ = $\{(\mathcal{D}^j_{{tr}}, \mathcal{D}^j_{{te}})\}^T_{j=0}$ \\
\STATE 4: \quad \quad \quad \quad \quad \; $\triangleright$ sample a pseudo incremental sequence\\
\STATE 5: \quad $\mathcal{P}$ = $\varnothing$ \quad \quad\quad $\triangleright$ empty cumulative pseudo test set\\
\STATE 6: \quad $\theta^{E,P} \leftarrow \theta^{E,P} -\gamma\nabla_{\theta^{E,P}}\mathcal{L}_{labeled}(\mathbf{x}^0_{tr},\mathbf{y}^0_{tr};\theta)$\\
\STATE 7: \quad \; $\triangleright$update parameters using pseudo labeled classes\\
\STATE 8: \quad  $\mathcal{P}$ = $\mathcal{P}\cup \mathcal{D}^0_{te}$ \; \quad 
 \quad $\triangleright$ accumulate test set of sess-$0$\\
\STATE 9: \quad \textbf{for} $j$ = $1,\cdots,T$ \textbf{do} \\
\STATE 10: \quad \quad $\tilde{\theta}^{E,P}$ = $\theta^{E,P}-\alpha\nabla_{\theta^{E,P}} \mathcal{L}_{soft}(\mathbf{x}^{j}_{tr};\theta)$\\ 
\STATE 11: \quad$\triangleright$ 
compute adapted params with unlabeled samples\\ 
\STATE 12: \quad\quad  $\mathcal{P}$ = $\mathcal{P}\cup \mathcal{D}^j_{te}$ \quad \quad $\triangleright$accumulate test set of sess-$j$\\ 
\STATE 13: \quad\quad $\theta \leftarrow \theta -\beta\nabla_{\theta}\sum_{(\mathcal{X},\mathcal{Y})\in \mathcal{P}} \mathcal{L}_{scl}(\mathcal{X},\mathcal{Y}; \tilde{\theta}^{E},\tilde{\theta}^P)$\\
\STATE 15: \quad\quad\quad\quad\quad\; $\triangleright$ update meta-params $\theta$ to new session\\
\STATE 16: \quad \textbf{end for}
\STATE 17: \textbf{end while}
\end{algorithmic}
\end{algorithm}

\noindent{\textbf{Meta-training.}} For each sampled sequence $\mathcal{D}$, we let the model continually explore the incoming unlabeled data in an unsupervised manner. To reduce the forgetting issue due to learning new knowledge, we utilize the bi-level optimization~\cite{finn2017model,chi2021test,chi2022metafscil} to directly formulate incrementally discovering without forgetting as the meta-objective. The meta-learning procedure is illustrated in Alg. \ref{Algorithm one} and Fig.~\ref{fig2}~(b). Concretely, we decouple the network as $\theta=\{\theta^{E}, \theta^{P}\}$, where $\theta^{E}$ and $\theta^{P}$ are the encoder and projection layers. At each incremental session, we aim to evaluate all the classes that have encountered so far. Hence, at the beginning of each sequence, we define an empty cumulative pseudo test set $\mathcal{P}$ to store the test samples from previous sessions. After that, we first train $\theta$ on the pseudo labeled classes ($j$ = $0$) using the unsupervised and supervised contrastive loss (Eq. \ref{jointly}). At each $j^{th}$ session ($j > 0$), we update $\theta$ on unlabeled samples $\mathcal{D}_{tr}^j = \{\mathbf{x}_{tr}^j\}$ via a few gradient steps:
\begin{equation}
\tilde{\theta}^{E,P} = \theta^{E,P}-\alpha\nabla_{\theta^{E,P}} \mathcal{L}_{soft}(\mathbf{x}_{tr}^j;\theta)
\label{eq:inner}
\end{equation}
where $\mathcal{L}_{soft}(\cdot)$ is the proposed soft neighborhood contrastive loss (Eq.~\ref{eq:soft}). It aims to discriminate uncorrelated samples while absorbing correlated ones. By thoroughly exploring the unlabeled data, it maintains comprehensive old knowledge while efficiently discovering novel classes.

Eq.~\ref{eq:inner} mimics how the model discovers novel classes on the incoming unlabeled data at test-time. Ideally, we like the adapted $\tilde{\theta}^{E,P}$ to perform well on all encountered classes. The test data from previous sessions and the current session separately reflect the catastrophic forgetting robustness and novel class discovery capability. Thus, we append $\mathcal{D}^j_{te}$ to $\mathcal{P}$. Accordingly, the meta-objective is defined as follows for the outer loop of the meta-level optimization: 
\begin{equation}
\min\limits_{\theta^E, \theta^P} \sum\nolimits_{(\mathcal{X},\mathcal{Y})\in \mathcal{P}} \mathcal{L}_{scl}(\mathcal{X},\mathcal{Y};\tilde{\theta}^E,\tilde{\theta}^P)
\label{meta-objective}
\end{equation}
where $\mathcal{L}_{scl}(\cdot)$ is the supervised contrastive loss in Eq.~\ref{uns_loss}. Note that the optimization is performed on $\theta$, although $\mathcal{L}_{scl}(\cdot)$ is a function of $\tilde{\theta}^{E}$ and $\tilde{\theta}^{P}$. The meta-objective in Eq. \ref{meta-objective} is then optimized using gradient descent, as shown in Line 13 of Alg. \ref{Algorithm one}. We empty $\mathcal{P}$ when all $T+1$ sessions are iterated. After meta-training, we obtain an initialization of the model $\theta$ which has been specifically trained to discover and learn novel objects from a sequence of unlabeled data.

\noindent{\textbf{Meta-testing.}}
It is worth mentioning that the procedure in Alg.~\ref{Algorithm one} aligns with the evaluation protocol. After discovering novel classes at each incremental session, the model is evaluated on all encountered classes. Our meta-objective optimizes the model towards what it is supposed to do at evaluation to maximize the performance. In addition, despite some uncertainties that may occur for unsupervised learning, the model is constrained by a fully supervised meta-objective. Thus, when training converges, the meta-model $\theta$ is ready to discover novel classes while maintaining the old knowledge by only running Line 10 of Alg. \ref{Algorithm one}. 

\section{Experiments}
\subsection{Dataset and setup}
\paragraph{Dataset.} We construct the C-GCD benchmark using three widely used datasets as in NCD~\cite{vaze2022generalized,roy2022class,zhong2021neighborhood}, \textit{i.e.}, CIFAR10 \cite{krizhevsky2009learning}, CIFAR100 \cite{krizhevsky2009learning} and Tiny-ImageNet \cite{le2015tiny}. 
Each dataset is split into two subsets, 1) large-scale labeled samples accounting for 80\% of the known classes data constitute a labeled set for offline training; and 2) the remaining data containing known and novel classes are used as an unlabeled set for continual object discovery. In Tab. \ref{dataset}, we summarize the dataset splits used in our training.  

\noindent\textbf{Session-wise data split.} All labeled samples $\mathcal{S}^0$ are used for offline training in our setting. During the online incremental learning stage, the unlabeled samples are dynamically added (\textit{i.e.}, sessions $t \geq 1$). Specifically, CIFAR10 is divided into 3 incremental sessions. In the $t^{th} (t\textgreater0)$ session, 3000 unlabeled images from 1 novel class and 2000 unlabeled images from $7+(t-1)\times 1$ known classes are added. CIFAR100 is divided into 4 sessions, in which 1500 unlabeled images from 5 novel classes and 2000 unlabeled images from $80+(t-1)\times5$ known classes are added in the $t^{th}$ session. The Tiny-ImageNet consists of 5 incremental sessions, each containing 3000 unlabeled images from 10 novel classes and 3000 unlabeled images from $150+(t-1)\times 10$ known classes. 

\noindent\textbf{Sequential task sampling.} During offline training, we use $\mathcal{S}^0$ to sample sequential tasks. We first split $\mathcal{S}^0$ into non-overlapping \textit{pseudo labeled} and \textit{novel} classes (4/3 for CIFAR10, 60/20 for CIFAR100 and 100/50 for Tiny-ImageNet). For each task, the \textit{pseudo labeled} set is first used to warm up the model , followed by $T$ incremental sessions of unlabeled samples containing the \textit{pseudo labeled} and \textit{novel} classes. Both the session number and the number of novel classes in each offline incremental session are consistent with the online incremental learning scenario.

\begin{table}[t]
\centering
\setlength{\tabcolsep}{2.0mm}{
\begin{tabular}{c|cc|cc}
\toprule[1pt]
\multirow{2}{*}{Dataset} &\multicolumn{2}{c|}{Labeled Set} &\multicolumn{2}{c}{Unlabeled Set} \\ \cline{2-5}
               & \#class &\#image &\#class &\#image \\\hline
CIFAR10  &7 &28000 &10 &22000   \\
CIFAR100 &80 &32000 &100 &18000  \\
Tiny-ImageNet &150 &60000 &200 &40000  \\
\bottomrule[1pt]
\end{tabular}}
\caption{ Datasets used in our experiments. We show the number
of classes in the labeled and unlabeled sets, as well
as the number of samples.}
\label{dataset}
\end{table}

\begin{table*}[t]
\centering
\small
\begin{subtable}[t]{1.0\linewidth}
\centering
\setlength{\tabcolsep}{2.36mm}{
\begin{tabular}{cccc|ccc|ccc|ccc}
\toprule[1pt]
\multirow{3}{*}{Methods} &\multicolumn{9}{c}{\textbf{CIFAR10} (Session Number)} &\multicolumn{3}{c}{Final} \\ \cline{2-10}                       
           &\multicolumn{3}{c}{1} &\multicolumn{3}{c}{2} &\multicolumn{3}{c}{3} &\multicolumn{3}{c}{Impro.}  \\ \cline{2-13} 
            &All &Old &New &All &Old &New &All &Old &New &All &Old &New\\\hline
RankStats &69.31 &70.20 &58.63 &65.23 &67.86 &51.20 &38.16 &50.01 &35.94 &+54.50 &+47.22 &+48.77\\
FRoST &73.92 &81.17 &66.45 &69.56 &79.73 &58.04 &67.73 &70.84 &51.13 &+24.93 &+26.39 &+33.58\\
VanillaGCD &89.24 &97.97 &81.80 &85.13 &96.67 &74.60 &86.41 &95.03 &76.75  &+6.25 &+2.20 &+7.96\\
GM &90.00 &98.41 &77.40 &87.39 &\textbf{99.01} &73.46 &87.86 &97.15 &78.93 &+4.80 &+0.08 &+5.78\\ \hline
MetaGCD(ours) &\textbf{95.38} &\textbf{99.07} &\textbf{89.15} &\textbf{93.34} &98.81 &\textbf{85.39} &\textbf{92.66} &\textbf{97.23} &\textbf{84.71} & & & \\
\bottomrule[1pt]
\end{tabular}}
\end{subtable}

\begin{subtable}[t]{1.0\linewidth}
\centering
\setlength{\tabcolsep}{1.26mm}{
\begin{tabular}{cccc|ccc|ccc|ccc|ccc}
\toprule[1pt]
\multirow{3}{*}{Methods} &\multicolumn{12}{c}{\textbf{CIFAR100} (Session Number)} &\multicolumn{3}{c}{Final} \\ \cline{2-13}                         
           &\multicolumn{3}{c}{1} &\multicolumn{3}{c}{2} &\multicolumn{3}{c}{3} &\multicolumn{3}{c}{4} &\multicolumn{3}{c}{Impro.}  \\ \cline{2-16} 
           &All &Old &New &All &Old &New &All &Old &New &All &Old &New &All &Old &New\\\hline
RankStats &62.33 &64.22 &31.60 &55.01 &58.55 &26.85 &51.77 &56.70 &25.47 &47.51 &54.59 &17.20 &+27.05 &+23.01 &+43.93 \\
FRoST &67.14 &68.57 &50.73 &67.01 &68.82 &52.60 &62.35 &65.48 &45.67 &55.84 &59.06 &42.95 &+18.72 &+18.54 &+18.18 \\
VanillaGCD &76.78 &77.91 &58.60 &73.67 &75.29 &60.70 &72.77 &74.72 &62.33 &71.44 &74.75 &58.20 &+3.12 &+2.85 &+2.93\\
GM &78.29 &\textbf{79.91} &66.00 &77.58 &\textbf{79.64} &61.13 &74.56 &77.60 &58.14 &72.02 &75.98 &56.32 &+2.54 &+1.62 &+4.81\\ \hline
MetaGCD(ours) &\textbf{78.96} &79.36 &\textbf{72.60} &\textbf{78.67} &79.41 & 66.81 &\textbf{76.06} &\textbf{78.20} &\textbf{64.87} 
&\textbf{74.56} &\textbf{77.60} &\textbf{61.13} 
\\
\bottomrule[1pt]
\end{tabular}}
\end{subtable}

\begin{subtable}[t]{1.0\linewidth}
\centering
\setlength{\tabcolsep}{0.5mm}{
\begin{tabular}{cccc|ccc|ccc|ccc|ccc|ccc}
\toprule[1pt]
\multirow{3}{*}{Methods} &\multicolumn{15}{c}{\textbf{Tiny-ImageNet} (Session Number)} &\multicolumn{3}{c}{Final} \\ \cline{2-16}                     
           &\multicolumn{3}{c}{1} &\multicolumn{3}{c}{2} &\multicolumn{3}{c}{3} &\multicolumn{3}{c}{4} &\multicolumn{3}{c}{5} &\multicolumn{3}{c}{Impro.}  \\ \cline{2-19} 
           &All &Old &New &All &Old &New &All &Old &New &All &Old &New &All &Old &New &All &Old &New\\\hline
RankStats &62.39 &64.54 &35.01 &55.89 &52.23 &34.20 &49.88 &46.17 &28.33 &44.20 &42.87 &24.50 &36.09 &35.20 &15.76 &+34.15 &+36.33 &+42.7\\
FRoST &64.92 &67.84 &46.28 &59.50 &61.86 &40.60 &57.86 &60.63 &39.14 &55.68 &59.71 &36.55 &50.49 &53.76 &33.37 &+19.75 &+17.77 &+29.09\\
VanillaGCD &75.92 &78.17 &62.15 &74.53 &77.73 &56.12 &73.64 &74.85 &57.31 &70.69 &71.13 &54.35 &66.15 &67.17 &54.43 &+4.09 &+4.36 &+4.03\\
GM &76.32 &\textbf{79.55} &63.60 &75.43 &78.10 &57.40 &72.63 &76.29 &54.80 &70.54 &\textbf{76.80} &51.50 &67.31 &\textbf{72.08} &50.90 &+2.93 &-0.55 &+7.56\\ \hline
MetaGCD(ours) &\textbf{78.67} &79.41 &\textbf{66.80} &\textbf{77.89} &\textbf{79.95} &\textbf{61.40} &\textbf{75.23} &\textbf{77.86} &\textbf{61.20} &\textbf{72.00} &75.61 &\textbf{57.55} &\textbf{70.24} &71.53 &\textbf{58.46} \\
\bottomrule[1pt]
\end{tabular}}
\end{subtable}
\caption{{\bf Performance (in \%) comparisons with the state-of-the-art methods on \textbf{CIFAR10}, \textbf{CIFAR100}, \textbf{Tiny-ImageNet} datasets}. The results of other methods are obtained by running their released codes under the C-GCD setting.}
\label{tab:2}
\end{table*}

\noindent{\textbf{Evaluation metrics.}} After learning the model on unlabeled samples at every online incremental stage, we follow~\cite{vaze2022generalized} to measure the clustering accuracy between the ground truth labels $y_i$ and the model's predictions $\hat{y}_i$ as:
\begin{equation}
ACC=\max\limits_{p\in \mathcal{P}(\mathcal{Y})}\frac{1}{N}\sum(\mathds{1}\{y_i = p(\hat{y}_i)\}),
\end{equation}
where $N$ is the total number of test samples and $\mathcal{P}(\mathcal{Y})$ is the set of all permutations of the class labels $\mathcal{Y}$ encountered so far. The optimal permutation can be obtained via the Hungarian algorithm \cite{kuhn1955hungarian}. Our main metric is $ACC$ on `All' classes, indicating the accuracy across all accumulated test sets so far. To decouple the evaluation on \textit{forgetting} and \textit{discovery}, we further report accuracy for both the `Old' classes subset (samples in the test set belonging to previous known classes) and `New' classes subset (samples in the test set belonging to novel classes). 

\noindent{\textbf{Implementation details.} }
Following \cite{vaze2022generalized}, we employ a vision transformer (ViT-B-16) \cite{dosovitskiy2021image} pretrained on ImageNet~\cite{deng2009imagenet} with DINO~\cite{caron2021emerging} as the feature extractor throughout the paper. We use the Adam optimizer and the learning rates in Alg.~\ref{Algorithm one} are set as $\gamma=0.1$, $\alpha=0.001$ and $\beta=0.0001$. We use a batch size of 256 and $\lambda=0.35 $ to balance the losses in Eq.~\ref{jointly}. Unless otherwise stated, we select the threshold $\epsilon$ to be 0.85 in Eq.~\ref{eq:nearest}. 
At the meta-training stage, we first perform training on \textit{pseudo labeled} set for 50 epochs, followed by 10 inner and 1 outer gradient updates for incremental sessions. At the meta-test stage, we directly perform 20 gradient updates to adapt using \textit{unlabeled} samples. Furthermore, we follow standard practice in self-supervised learning to use the same projection head as in \cite{caron2021emerging} and discard it at test-time. 

\begin{table*}[htbp]
\begin{center}
\setlength{\tabcolsep}{1.0mm}{
\begin{tabular}{c|ccc|ccc|ccc|ccc|ccc}
\toprule[1pt]
\multirow{3}{*}{Methods} &\multicolumn{12}{c|}{\textbf{CIFAR100} (Session Number)}  &\multicolumn{3}{c}{Average} \\ \cline{2-13}
                         &\multicolumn{3}{c|}{1} &\multicolumn{3}{c|}{2} &\multicolumn{3}{c|}{3}  &\multicolumn{3}{c|}{4} &\multicolumn{3}{c}{Acc}\\ 
                         &All &Old &New &All &Old &New &All &Old &New &All &Old &New &$m$A &$m$O &$m$N \\ \hline
Baseline &76.78 &77.91 &58.60 &73.67 &75.29 &60.70 &72.77 &74.62 &62.33 &71.44 &74.75 &58.20 &73.67 &75.64 &59.96 \\
+$\mathcal{L}_{CN}$ &77.29 &78.53 &65.44 &75.20 &77.67 &63.08 &74.55 &75.91 & 63.63 &72.62 &75.43 &59.40 &74.92 &76.89 &62.89 \\
+$\mathcal{L}_{SP}$ &77.92 &78.45 &68.78 &76.53 &78.22 &64.71 &74.86 &77.24 &64.09 &73.34 &76.87 &60.60 &75.66 &77.70 &64.55\\
+ Meta-learning &78.96 &79.36 &72.60 &78.67 &79.41 & 66.81 &76.06 &78.20 &64.87  &74.56 &77.60 &61.13 &77.06 &78.64 &66.35 \\
\bottomrule[1pt]
\end{tabular}}
\caption{{\bf Ablation study of various components of our MetaGCD on the CIFAR100 dataset}. We report `All', `Old' and `New' class accuracy for each incremental session, and the average of all sessions such as mean `All' ($mA$), mean `Old' ($mO$) and mean `New' accuracy ($mN$). Here \textbf{CN} denotes candidate neighbors and \textbf{SP} denotes soft positiveness. }
\label{ablation_study}
\end{center}
\end{table*}

\subsection{Comparison with the state-of-the-art}
Since this paper considers a new problem setting, there is no prior work that we can directly compare. Nevertheless, we choose SOTA methods on NCD and run their codes under our C-GCD setting, 
including RankStats \cite{han2021autonovel}, VanillaGCD \cite{vaze2022generalized}, and recent continual NCD models FRoST \cite{roy2022class}, GM \cite{zhang2022grow}. Both RankStats and FRoST train two classifiers on top of a shared feature representation. The first head is fed images from the labeled set and is trained with the cross-entropy loss, while the second head sees only images from unlabeled images of novel classes. In order to adapt RankStats and FRoST to C-GCD, we train them with a single classification head for the total number of classes in the dataset. The sequential version of VanillaGCD is adopted and serves as the baseline for our model. We leverage the original training mechanism for GM.

\begin{figure}[t]
\centering
\centerline{\includegraphics[scale=0.40]{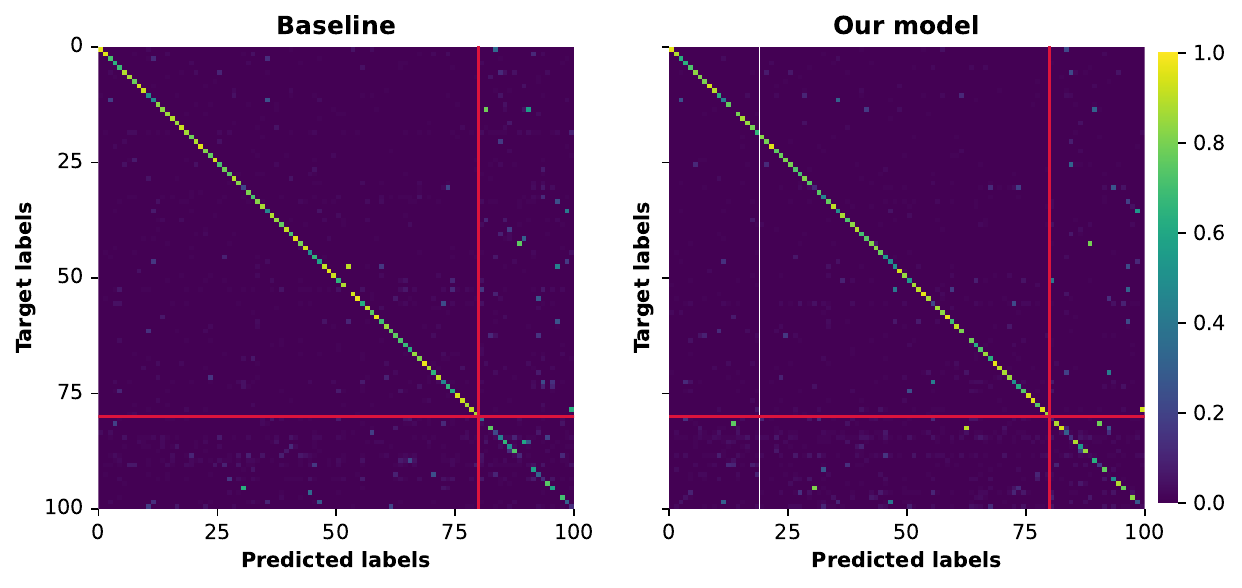}}
\caption{{\bf Class-wise performance on the CIFAR100 dataset}. The confusion matrices show that our model significantly improves the baseline for both \textit{known} and \textit{novel} classes (separated by the red line). Especially for novel classes, the confusion matrix of our method has more concentrated values along the diagonal.} 
\label{fig3}
\end{figure}

In Tab.~\ref{tab:2}, we report the \textit{All}/\textit{Old}/\textit{New} class accuracy per incremental session for all methods, and the relative improvement for the final session. As we can see, the proposed method consistently outperforms all other methods on all three datasets among the most incremental sessions. Specifically, our MetaGCD surpasses the most recent method GM by 5.78\%, 4.81\% and 7.56\% on CIFAR10, CIFAR100 and Tiny-ImageNet datasets for the final \textit{New} classes accuracy. Besides, our model outperforms the baseline VanillaGCD by 6.25\%, 3.12\% and 4.09\% for the final \textit{All} classes. We also report the class-wise performance via the confusion matrices shown in Fig.~\ref{fig3}. It is obvious that the baseline performs poorly, especially on the \textit{novel} classes. However, the proposed model has a significant gain in discovering \textit{novel} classes. Moreover, less forgetting is observed in our method, as more values are concentrated on the diagonal. 

\subsection{Ablation Study} 
We conduct ablation studies on the CIFAR100 dataset to evaluate each component in our proposed framework. 

\begin{figure}[t]
\centering
\centerline{\includegraphics[scale=0.19]{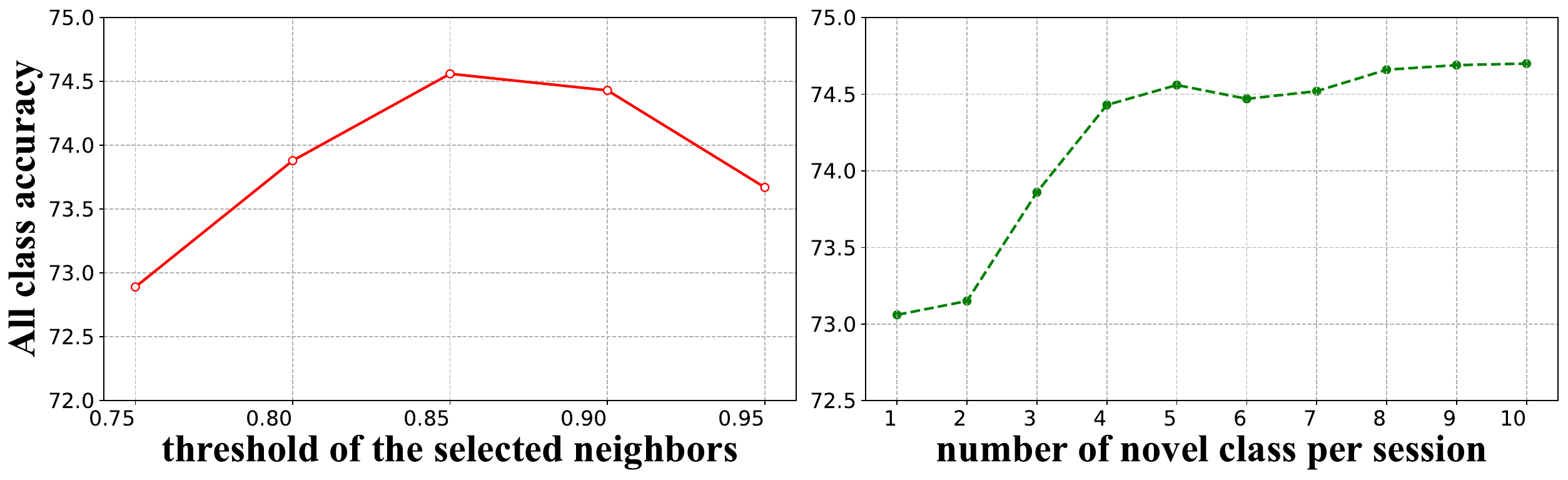}}
\caption{{\bf Hyper-parameter analysis on the CIFAR100 dataset regarding feature similarity threshold (left) and various numbers of novel classes (right)}. An appropriate threshold or the number of new classes helps to stabilize the training process and improve performance.} 
\label{fig4}
\end{figure}

\noindent\textbf{Importance of neighborhood.} In our contrastive learning framework, we compute the soft correlation to allow more positive pairs to contribute to the loss. As reported in the second row of Tab.~\ref{ablation_study}, considering the neighborhood achieves a performance gain of 1.25\% (\textit{i.e.}, 74.92\% \textit{v.s.} 73.67\% on the mean accuracy of \textit{All} classes). The performance gain may come from the abundant comparisons from positive samples, which facilitates the current instance to align with more highly-correlated samples. We also conduct experiments to assess the sensitivity of threshold $\epsilon$ when selecting the positive neighbor instances. Increasing $\epsilon$ allows more strict positive pairs, but some partially related samples might be ignored. On the other hand, reducing $\epsilon$ increases the likelihood of introducing true negative samples, which may negatively impact model performance. As empirically found in the left side of Fig.~\ref{fig4}, a trade-off should be made, and a threshold of 0.85 achieves the best performance.

\noindent\textbf{Importance of soft positiveness.} We then analyze the importance of soft positiveness to the recognition performance in the third row of Tab.~\ref{ablation_study}. When we compute a correlation weight for each selected neighbor, the clustering accuracy on \textit{Novel} classes increases from 62.89\% to 64.55\%. It indicates that the binary labeling strategy is insufficient to measure the correlation at the feature space, thus causing the backbone network to produce less discriminative representations compared with soft labeling methods. In the lower part of Fig.~\ref{fig2} (a), we show the correlation weight of selected neighbors with an input instance. The high score corresponds to the same category while the low score corresponds to less-correlated categories, which shows that our attention module is effective in modeling correlations between the input instance and each candidate neighbor.

\noindent\textbf{Effectiveness of meta-learning.} Our meta-learning optimization strategy further improves \textit{All} classes accuracy to 74.56\% for the final incremental session in the last row of Tab.~\ref{ablation_study}. It demonstrates the effectiveness of the proposed method where the meta-objective specifically forces the model to discover novel classes without forgetting old classes. Additionally, we analyze the impact of the number of novel classes that are sampled during meta-training. To investigate this, we train separate models by setting the number of novel classes in the range of \{1, 10\}. As illustrated on the right side of Fig.~\ref{fig4}, a larger number of classes for sequence tasks is more optimal. When there are fewer classes, the model is at a higher risk of overfitting to certain classes rather than learning how to incrementally learn. 

\section{Conclusion}

In this paper, we propose a more realistic setting for real-world applications, namely C-GCD. The ultimate goal of C-GCD is to discover novel classes while keeping the old knowledge without forgetting. We propose a meta-learning based optimization strategy to directly optimize the network to learn how to incrementally discover with less forgetting. In addition, we introduce a soft neighborhood contrastive learning to utilize the soft positiveness to adaptively support the current instances from their neighbors. Extensive experiments on three datasets demonstrate the superiority of our method over state-of-the-art methods.

\section{Acknowledgments}
This work was supported by the Fundamental Research Funds for the Central Universities (No. 2022JBZY019), the National Key Research and Development Project (No. 2018AAA0100300) and an NSERC Discovery grant.


\end{document}